\documentclass{article}
\usepackage{spconf,amsmath,graphicx}

\usepackage{caption} 
\usepackage{algorithm}
\usepackage{algorithmic}
\usepackage{booktabs}
\usepackage{multirow}
\usepackage{amsmath}
\usepackage{amssymb}
\usepackage{mathtools}
\usepackage{amsthm}
\usepackage{stfloats}

\usepackage[caption=false,font=footnotesize,captionskip=0.1mm,farskip=1mm]{subfig}

\title{Thundernna: a white box adversarial attack}
\name{Linfeng Ye$^*$, Shayan~Mohajer~Hamidi$^*$\thanks{$^*$ Authors contributed equally.}}
\address{University of Waterloo\\Dept. of Electrical and Computer Engineering\\Waterloo, ON N2L 3G1}

\begin{document}
	\maketitle
	\section{Abstract}

The existing literature highlights the vulnerability of neural networks trained using naive gradient-based optimization methods to adversarial attacks. The injection of small malicious perturbations into ordinary input is sufficient to mislead the neural network. Simultaneously, countering adversarial attacks is pivotal for enhancing the network's robustness. Training against adversarial examples can fortify neural networks against certain types of adversarial attacks. Additionally, adversarial attacks on neural networks can unveil the intrinsic characteristics of these networks, which function as complex high-dimensional non-linear entities, as discussed in \cite{2019Second,2014Intriguing}. In this paper, we introduce a first-order method to attack neural networks. Compared to other first-order attacks, our method boasts a significantly higher success rate. Furthermore, it outpaces second-order attacks and multi-step first-order attacks in terms of speed.

	\section{Introduction}
An adversarial attack involves a malicious attempt to perturb a data point \(x_0 \in \mathbb{R}^d\) into another point \(x \in \mathbb{R}^d\), such that \(x\) falls within a specific target adversarial class. In the context of an image classification task, for instance, if \(x_0\) represents an image of a cat, an adversarial attack aims to subtly modify the image \(x_0\) in order to induce misclassification by the classifier.

	Geoffrey Hinton proposed the idea of deep learning in 2006. Benefiting from the emergence of big data and large-scale parallel computing, deep learning has become one of the most active studies of computer science. The multi-layer and non-linear structure equips it with strong feature expression capability and modelling capability towards complicated tasks. In recent years, the development of deep learning has brought up a series of research, especially in the image recognition area. Experiments on some standard test sets indicate the recognition ability of deep learning models has already reached a human-like level. However, people still question whether for an abnormal input, whether deep learning model can generate satisfactory results. 
	Szegedy et al. [1] brought up the idea of adversarial examples in the research paper published in ICLR2014, which is, the input samples that formed by intentionally adding minor perturbation to the data set will generate a false output with high certainty after disturbed. Their research suggests, in most cases, the model that has different structures gained by training on various subsets of the training set will mistakenly distinguish the same adversarial example, which means that the adversarial examples become a blind spot of the training algorithm. Nguyen et al. \cite{2015fool} found that for fooling examples that humans cannot distinguish at all, deep learning models will classify them with high certainty, Kdnuggets points out that the fragility of deep learning in terms of adversarial examples is not uniquely owned by deep learning, but commonly exist in many machine learning models. Therefore, further research in adversarial examples is beneficial to the progress of entire machine learning and deep learning areas. 
	Nowadays there are multiple methods to calculate adversarial examples. Akhter et al. [3] concluded over 12 kinds of attack methods that can fool classify models. Furthermore, besides researching attacking classification/distinguishment in computer vision, researchers also are doing research on attacks towards other areas and directions. For example, attacks towards autoencoders, generative models, semantic segmentation and object detection. Besides understanding adversarial examples’ existence space in mathematical areas, many researchers are trying to understand adversarial attacks being added to physical objects in the real world. For example, Athalye et al. \cite{2018Synthesizing} indicate we can even generate a 3-D printed physical object adversarial example to fool deep neural network classifier. Gu et al. \cite{2017Universal}\cite{2018Synthesizing} discuss interesting work: fooling neural network in street signs. The neural network recognizes the stop sign as 45 speed limits. 
	In the mathematical world, more work focuses on generating perturbations that lead to particular image inputs being classified falsely. However, it has been proved that we can generate image-unrelated adversarial samples. Moosavi-Dezfooli et al. \cite{2017Fast} indicate that given a target model and data set we can calculate a single perturbation. Mopuri et al. demonstrate their algorithm (FFF\cite{2017Fast}, GDUAp\cite{2017Generalizable}) to generate image-unrelated perturbation. These perturbations can deceive the target model without knowing the data distribution. They have proven that their meticulously designed perturbation can migrate to three different computer vision tasks, including classification, deep estimation and division.
	Generating adversarial examples usually takes a large amount of time and computing power. For anti-attack training, we need huge adversarial examples to find potential safety problems. This brings us to the focus on accelerating the generation of adversarial examples. 
	\section{Related Work}
	Many work \cite{2015Explaining}\cite{2015Deep}\cite{2016robustness}\cite{2018Analysis}\cite{2016Adversarial}\cite{2016limitations}\cite{2017One}\cite{2017Towards}\cite{2016Deepfool} research on adversarial perturbation to fool image classifier. Szegedy et al. \cite{2014Intriguing} first introduce the concept of adversarial examples and describe adversarial perturbation generation as an optimization problem. Goodfellow et al. [21] brought up an optimized maximum norm constraint perturbation method, call it “Fast Gradient Sign Method”(FGSM), to improve the computation efficiency. Kurakin et al. \cite{2016Adversarialexamples} brought up a “Fundamental iterative method”, which use FGSM iteratively to generate perturbation. Papernot et al. \cite{2016limitations} constructed an adversarial indication map to point out the ideal place for effective influence. Moosavi et al.’s DeepFool \cite{2016Deepfool} further improves the efficiency of adversarial perturbation. Moosavi-Dezfooli et al. [29] found that the image classifier has image-unrelated adversarial perturbation. Similarly, Metzen et al. \cite{2017Universaladversarial} brought up UAP for semantic segmentation task. They expanded Kurakin et al.’s iterative FGSM attak[31] to change every pixel’s prediction label. Mopuri et al. found general perturbation for data independence, not to sample any data from data distribution. They brought up a new non-data target algorithm to generate general adversarial perturbation, called it “FFF” \cite{2017Fast}. Their afterward work GDUAP \cite{2017Generalizable} improved adversarial attack’s effect and proved the viability of their method in computer vision tasks. 
	In order to solve linear assumption problem in FGSM, paper \cite{2014Threat} mentions using projected gradient descent (PGD) approach to solve the maximum value problem. PGD is an iterative attack, compared to FGSM which only has one iteration, PGD has multiple iterations. Each time a small step, each iteration will project perturbation into a specified range. 
	Since each time only moves a small step, the partial linear assumption is basically valid. After multiple steps, we can reach the optimum solution, which is the strongest attack result. The thesis also proved that the attack samples gained from PGD algorithm is the strongest in first-order adversarial examples. The first-order adversarial examples mean according to the first-order gradient adversarial example. If the model is robust to the samples generated by PGD, it is basically robust to all first-order adversarial examples. The experiment indicates that the model trained by PGD algorithm has high robustness. 
	Although PGD is simple and effective, there is a problem that it doesn’t have high computational efficiency. Without using PGD, m times iteration only has m gradient calculation. However, for PGD, every time the gradient decreases needs corresponding k steps gradient increases. Therefore, compared to methods with no adversarial training, PGD needs m(k+1) times gradient calculation.
	\section{Contributions}
	In this project, we developed a new first-order adversarial attack method. It uses the second-order algorithm to attack the integral of the neural network instead of the attack neural network itself. At the same time, it has the same time complexity as the first-order attack. We implement the algorithm and successfully attack resnet18 with pre-trained parameters.  which imply the integral of the neural network still has the characteristics of a convex function.
	Contrasting with previous work, we attempt to develop one step white box attacker, with the lower computational load which attacks instead of the network itself
	Further we ensemble the model by TVM frame work to accelerate the speed up the generation of adversarial examples.
	\section{Methods }
		In this paper, we developed a novel one-step first-order white box adversarial attack, which is much faster than a PGD attack and more powerful than FGSM.
		
		First, As we discussed in the previous section, an adversarial attack could easily treat the model train on the data generated by the first-order attack, which implies that a second-order attack is more powerful than a first-order attack, it's quite intuitive, In the training process, The model use second-order optimizer like Newton method to train would converge in fewer epochs. However, in all second-order optimizer/adversarial attacks Hessian matrix is required, which is so time-consuming that makes these second-order methods unpopular in both industry and academia.\\
		In Newton's method, We want to minimize a convex objective function, say $f(x)$ over x.
		Do first-order Taylor's expansion on f(x),
		$$f(x)\approx f(x_0)+f'(x_0)(x-x_0)+\frac{1}{2}f''(x)(x-x_0)^2$$
		To find the minima of the function $f(x)$, we take the derivative on both sides and set it equal to zero,
		$$f'(x) = f'(x_0) + f''(x_0)(x-x_0)=0$$
		$$f'(x) = f'(x_0) + f''(x_0)(x-x_0)=0$$
		$$x = x_0-\frac{f'(x_0)}{f''(x_0)}$$
		If we want to do an adversarial attack, we could simply change the minus to plus.\\ Then, If we let $f'(x)$ instead of $f(x)$ be our model and loss function, the function we optimized in this scenario is $f(x)=\int loss(model(x)) dx$. That means the function we use should have several 
		properties, first, when the loss is equal to zero when it arrives at the minima. Further, The $f(x)=\int loss(model(x)) dx$ should be a convex function. The rest of the problem is to find a loss function which meets all the requirements. We found that the negative log-likelihood loss with the softmax function meets all the requests. 
		
	First, The negative log-likelihood loss equals zero on perfect input(1), further, its integration is a convex function on the domain of $\left\{0 \rightarrow\infty\right\}$.
	$$\lim\limits_{\mu\rightarrow 0^+}\int_{\mu}^{t}-log(x)dx$$
	$$=-\lim\limits_{\mu\rightarrow 0^+}[xlog(x)|_\mu^t-\int_{t}^{\mu}xd(log(x))]$$
	$$=-\lim\limits_{\mu\rightarrow 0^+}[tlog(t)-\mu\log(\mu)-t+\mu]$$
	$$=tlog(t)-t$$
	\begin{figure}
		\centering
		\includegraphics[width=\columnwidth]{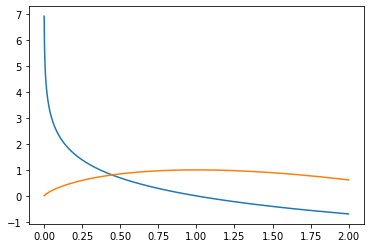}
	\end{figure}
	In this project, we use it as the loss function. As we proofed and the figure shows, The integral of the loss function is convex. 
	
	When it comes to our project, we assume that the input image is independent of the added noise. Thus, the derivative of the input image and the noise to the image is 1. Therefore, the molecule can be approximated as all 1's matrix, and the subsequent experiments also confirm the correctness of our conjecture. Finally, the the adversarial noise we get is  $$adversarial\ example = image + \frac{\overrightarrow{1}}{\frac{d\ nll\_loss}{ d\ image}}$$\\
	At the same time, from another point of view, we can find that the denominator of adversarial noise is just the loss function itself. We are assuming that the network we attack has completed training and the parameters have converged. At this time, the loss of the network can be approximated as a constant for how to effectively the loss function and input. At this time, it is assumed that each pixel is independent of the other. We can easily draw the above conclusion.\\
	Notice that, in the proof, there's a bug that the neural network is not a convex function, however, in real life, gradient-based optimizer works well. And the experiment shows that our attack is effective.
	\section{Experimental Results}
	We tested our algorithm on the Imagenet 2012 validation set, which contains 50000 pictures and their corresponding labels. using resnet18. The following picture shows the adversarial example generated by Thunderna.
	  Then we compare its performance with FGSM\cite{2015Explaining} PGD\cite{2017Towards} and second-order adversarial attack \cite{2019Second} under the same disturbances budget. Then we show successful adversarial examples generated by various algorithms. 
	\begin{figure}[ht]
 \subfloat[]{\includegraphics[width=0.2\columnwidth]{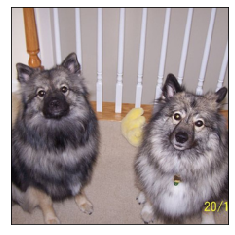}\label{fig1}}
	\subfloat[]{\includegraphics[width=0.2\columnwidth]{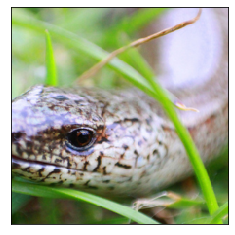}\label{fig2}}
 \subfloat[]{\includegraphics[width=0.2\columnwidth]{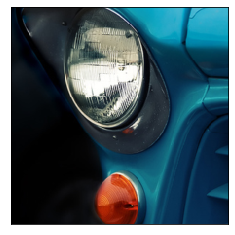}\label{fig3}}
	\subfloat[]{\includegraphics[width=0.2\columnwidth]{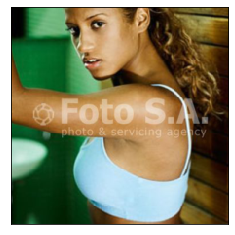}\label{fig4}} 
	\subfloat[]{\includegraphics[width=0.2\columnwidth]{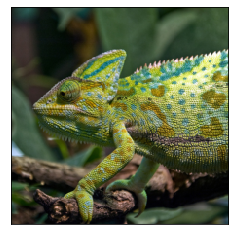}\label{fig5}} \\

 \subfloat[]{\includegraphics[width=0.2\columnwidth]{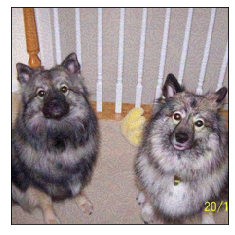}\label{fig1}}
	\subfloat[]{\includegraphics[width=0.2\columnwidth]{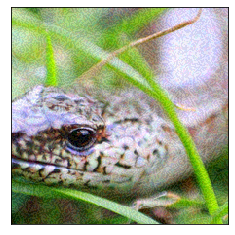}\label{fig2}}
 \subfloat[]{\includegraphics[width=0.2\columnwidth]{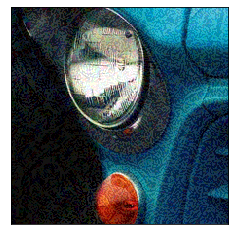}\label{fig1}}
	\subfloat[]{\includegraphics[width=0.2\columnwidth]{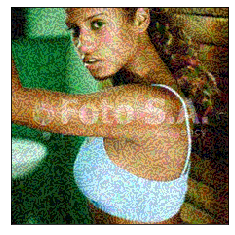}\label{fig2}} 
	\subfloat[]{\includegraphics[width=0.2\columnwidth]{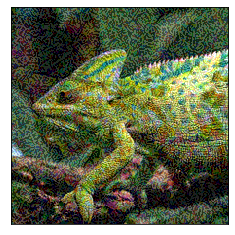}\label{fig2}}

		\caption{Thundernna attack examples and original images with budget 0.1 to 0.5.}
	\end{figure}

	\begin{figure}[ht]
 \subfloat[]{\includegraphics[width=0.2\columnwidth]{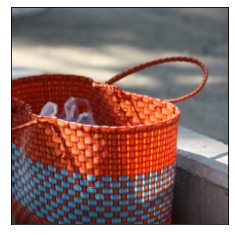}\label{fig1}}
	\subfloat[]{\includegraphics[width=0.2\columnwidth]{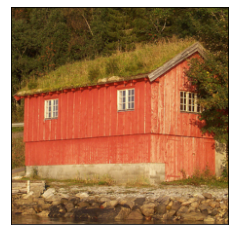}\label{fig2}}
 \subfloat[]{\includegraphics[width=0.2\columnwidth]{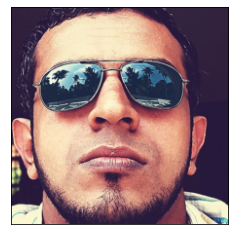}\label{fig3}}
	\subfloat[]{\includegraphics[width=0.2\columnwidth]{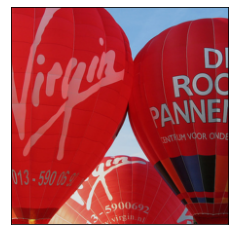}\label{fig4}} 
	\subfloat[]{\includegraphics[width=0.2\columnwidth]{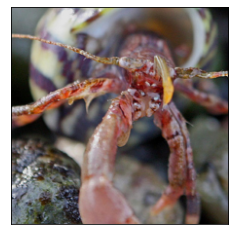}\label{fig5}} \\

 \subfloat[]{\includegraphics[width=0.2\columnwidth]{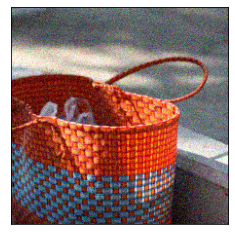}\label{fig1}}
	\subfloat[]{\includegraphics[width=0.2\columnwidth]{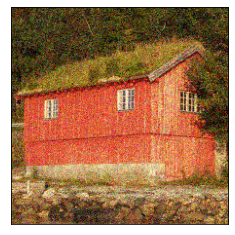}\label{fig2}}
 \subfloat[]{\includegraphics[width=0.2\columnwidth]{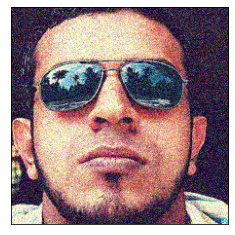}\label{fig1}}
	\subfloat[]{\includegraphics[width=0.2\columnwidth]{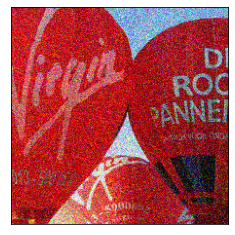}\label{fig2}} 
	\subfloat[]{\includegraphics[width=0.2\columnwidth]{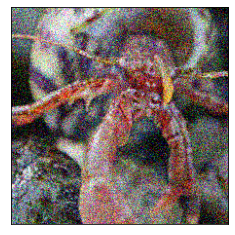}\label{fig2}}

		\caption{FGSM attack examples and original images with budget 0.1 to 0.5.}
	\end{figure}

	\begin{figure}[ht]
 \subfloat[]{\includegraphics[width=0.2\columnwidth]{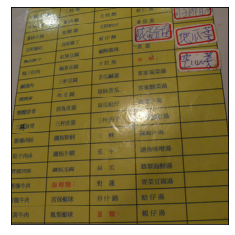}\label{fig1}}
	\subfloat[]{\includegraphics[width=0.2\columnwidth]{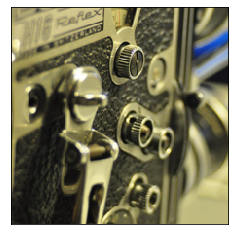}\label{fig2}}
 \subfloat[]{\includegraphics[width=0.2\columnwidth]{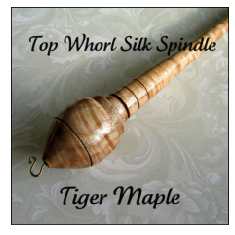}\label{fig3}}
	\subfloat[]{\includegraphics[width=0.2\columnwidth]{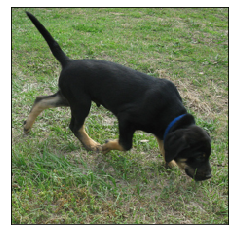}\label{fig4}} 
	\subfloat[]{\includegraphics[width=0.2\columnwidth]{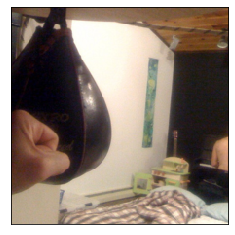}\label{fig5}} \\

 \subfloat[]{\includegraphics[width=0.2\columnwidth]{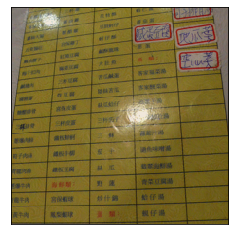}\label{fig1}}
	\subfloat[]{\includegraphics[width=0.2\columnwidth]{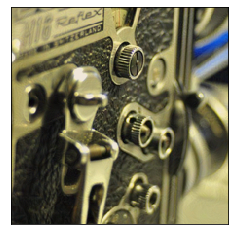}\label{fig2}}
 \subfloat[]{\includegraphics[width=0.2\columnwidth]{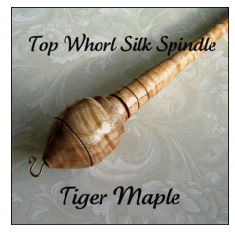}\label{fig1}}
	\subfloat[]{\includegraphics[width=0.2\columnwidth]{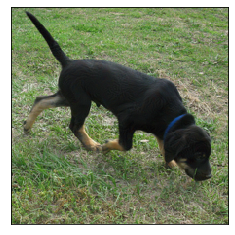}\label{fig2}} 
	\subfloat[]{\includegraphics[width=0.2\columnwidth]{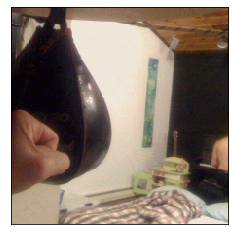}\label{fig2}}

		\caption{PGD attack examples and original images with budget 0.1 to 0.5.}
	\end{figure}

 	\begin{figure}[ht]
 \subfloat[]{\includegraphics[width=0.2\columnwidth]{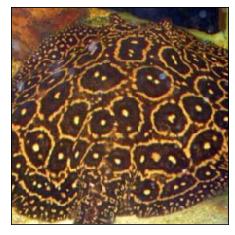}\label{fig1}}
	\subfloat[]{\includegraphics[width=0.2\columnwidth]{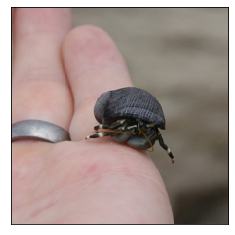}\label{fig2}}
 \subfloat[]{\includegraphics[width=0.2\columnwidth]{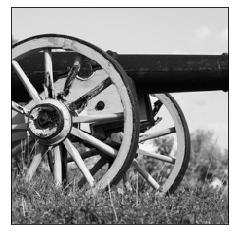}\label{fig3}}
	\subfloat[]{\includegraphics[width=0.2\columnwidth]{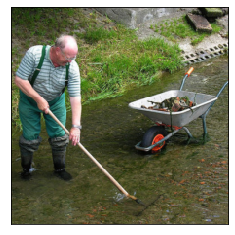}\label{fig4}} 
	\subfloat[]{\includegraphics[width=0.2\columnwidth]{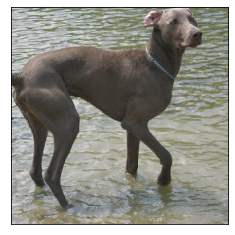}\label{fig5}} \\

 \subfloat[]{\includegraphics[width=0.2\columnwidth]{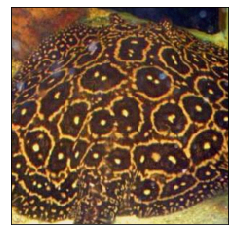}\label{fig1}}
	\subfloat[]{\includegraphics[width=0.2\columnwidth]{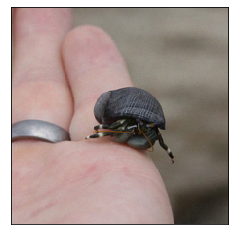}\label{fig2}}
 \subfloat[]{\includegraphics[width=0.2\columnwidth]{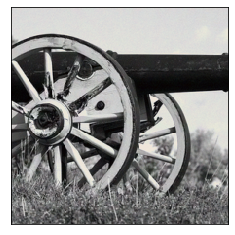}\label{fig1}}
	\subfloat[]{\includegraphics[width=0.2\columnwidth]{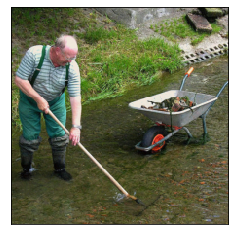}\label{fig2}} 
	\subfloat[]{\includegraphics[width=0.2\columnwidth]{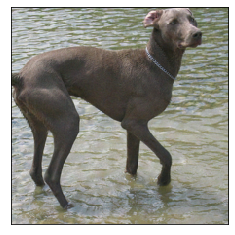}\label{fig2}}

		\caption{Second-order attack examples and original images with budget 0.1 to 0.5.}
	\end{figure}

	\newpage
	We can find that the loop-based algorithm can generate adversarial examples with much less noise than the one-step algorithm under the same budget. At the same time, the success rate is much higher than that of the one-step algorithm. Still, because the multi-step attack needs to be based on the picture generated by the last attack before the next attempt, it is impossible to optimize and eliminate the time sequence. The time required to generate the adversarial examples is 8 to 9 times that of the one-step attack.
	
\begin{table}[]
    \resizebox{0.99\linewidth}{!}{
    \begin{tabular}{l|llll} 
        \hline
        Methhods & Thundernna      & FGSM            & PGD              & second order attack \\ \hline
        Time     & 0.19s/50 & 0.16s/50  & 0.909s/50  & 0.955s/50     \\ \hline
    \end{tabular} 
    }
\end{table}
\begin{figure}[ht]
    \centering
    \includegraphics[width=\columnwidth]{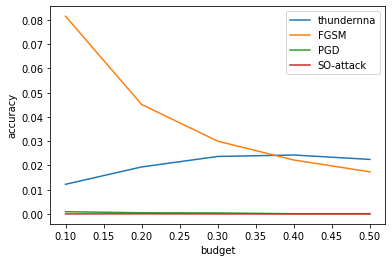}
\end{figure}	
	
	\section{Conclusions}
	
	We have developed a neural network confrontation algorithm. At the same time, we try to use the TVM framework to accelerate the process of forward and backward propagation of the neural network to accelerate the speed of the algorithm to generate confrontation examples.\\
	Under the same budget, our algorithm has a higher attack success rate than FGSM, which is also a first-order adversarial attack. However, the time complexity of the attack did not increase during the actual test. However, because it is a one-step algorithm and focuses on speed, its success rate is lower than that of multi-step algorithms, such as second-order attacks or PGD, under all budgets.
	
	\section{Future Work}
	In the future, we attempt to conduct adversarial training based on our attack algorithm to improve the robustness of the neural network and change the algorithm from a one-step attack to multiple-step attacks within loops to improve the success rate of attacks. At the same time, we noticed a strange phenomenon in the experiment. When the budget increases, the success rate of our algorithm attack decreases.  It is contrary to our intuition. We hope to continue exploring this attack's characteristics and look forward to discovering other neural networks. Interesting properties.

\bibliographystyle{IEEEbib}
\bibliography{strings,refs}

\end{document}